\documentclass[10pt,twocolumn,letterpaper]{article}

\usepackage[pagenumbers]{cvpr} 

\usepackage{float}
\usepackage{amssymb} 
\usepackage{multirow} 
\usepackage{graphicx} 
\usepackage{array}    
\usepackage{pifont}
\usepackage{bbding}
\usepackage{makecell}
\usepackage[skip=4pt]{caption} 
\definecolor{cvprblue}{rgb}{0.21,0.49,0.74}
\usepackage[pagebackref,breaklinks,colorlinks,allcolors=cvprblue]{hyperref}

\def\paperID{*****} 
\def\confName{CVPR}
\def\confYear{2026}
\title{EchoAgent: Towards Reliable Echocardiography Interpretation with ``Eyes'', ``Hands'' and ``Minds''}
\author{
Qin Wang\textsuperscript{1}, 
Zhiqing He\textsuperscript{1},
Yu Liu\textsuperscript{2}, 
Bowen Guo\textsuperscript{1},
Zeju Li\textsuperscript{1},
Miao Zhao\textsuperscript{3},
Wenhao Ju\textsuperscript{4},
Zhiling Luo\textsuperscript{3},\\
Xianhong Shu\textsuperscript{2 *},
Yi Guo\textsuperscript{1 *}, 
Yuanyuan Wang\textsuperscript{1}
\thanks{Corresponding authors}\\
\textsuperscript{1}Fudan University
\textsuperscript{2}Fudan Zhongshan Hospital
\textsuperscript{3}Fuwai Yunnan Hospital
\textsuperscript{4}Fuwai Beijing Hospital\\
}

\begin{document}
\maketitle
\begin{abstract}
Reliable interpretation of echocardiography (Echo) is crucial for assessing cardiac function, which demands clinicians to synchronously orchestrate multiple capabilities, including visual observation (``eyes''), manual measurement (``hands''), and expert knowledge learning and reasoning (``minds''). While current task-specific deep-learning approaches and multimodal large language models have demonstrated promise in assisting Echo analysis through automated segmentation or reasoning, they remain focused on restricted skills, i.e., ``eyes-hands'' or ``eyes‑minds'', thereby limiting clinical reliability and utility. To address these issues, we propose EchoAgent, an agentic system tailored for end‑to‑end Echo interpretation, which achieves a fully coordinated ``eyes‑hands‑minds'' workflow that learns, observes, operates, and reasons like a cardiac sonographer. First, we introduce an expertise‑driven cognition engine where our agent can automatically assimilate credible Echo guidelines into a structured knowledge base, thus constructing an Echo-customized ``mind''. Second, we devise a hierarchical collaboration toolkit to endow EchoAgent with ``eyes-hands'', which can automatically parse Echo video streams, identify cardiac views, perform anatomical segmentation, and quantitative measurement. Third, we integrate the perceived multimodal evidence with the exclusive knowledge base into an orchestrated reasoning hub to conduct explainable inferences. We evaluate EchoAgent on CAMUS and MIMIC‑EchoQA datasets, which cover 48 distinct echocardiographic views spanning 14 cardiac anatomical regions. Experimental results show that EchoAgent achieves state‑of‑the‑art performance across diverse cardiac structure function analyses, yielding overall accuracy scores of up to 80.00\%. Importantly, EchoAgent empowers a single system with abilities to learn, observe, operate and reason like a cardiac sonographer, which holds great promise for delivering reliable and clinically‑actionable Echo interpretation.
\end{abstract}

%
\vspace{-2ex}
\section{Introduction}
\begin{figure}[htb]
\includegraphics[width=\columnwidth]{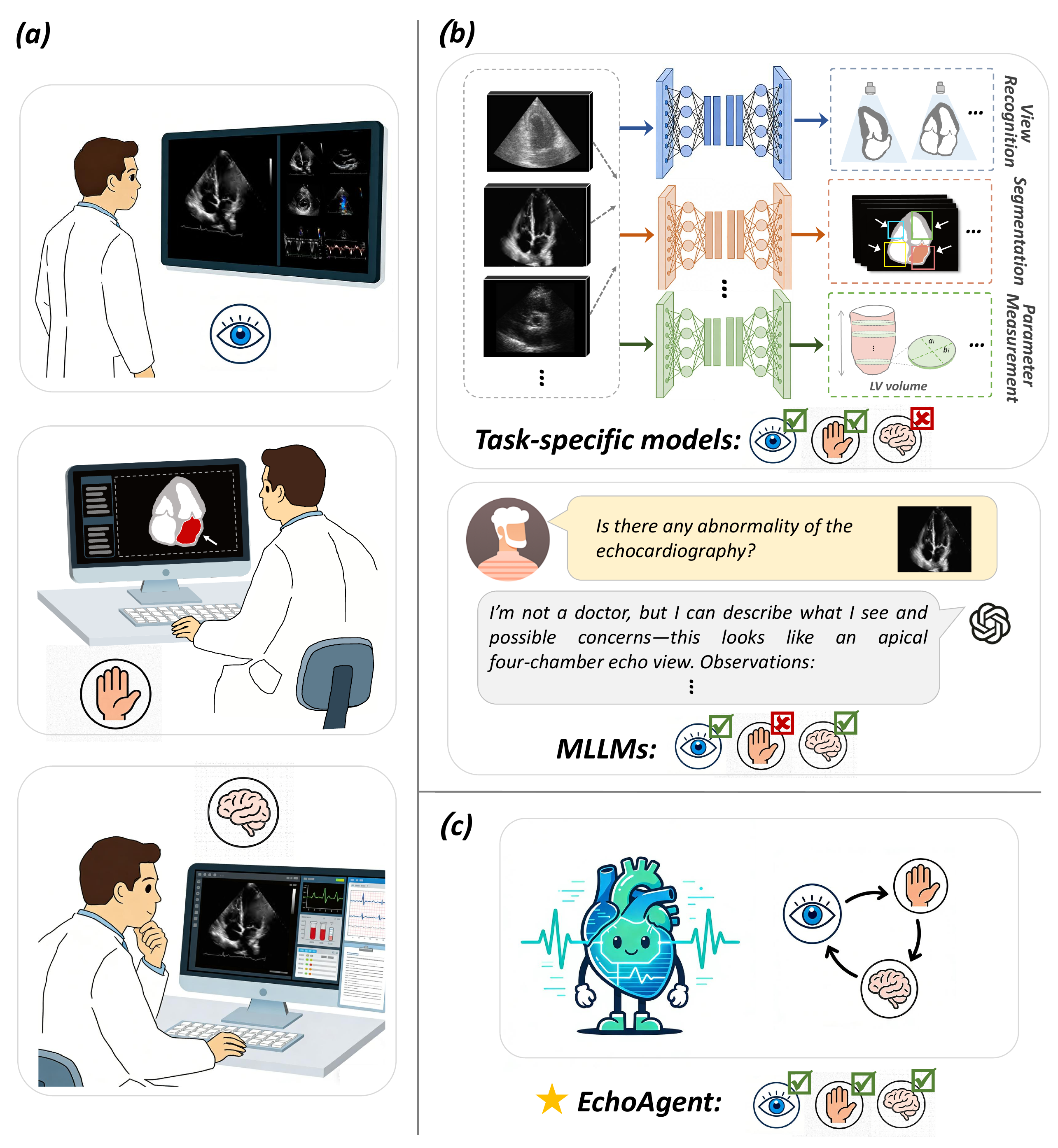}
\caption{Motivations and challenges of Echo interpretation. Motivations: (a) Schematic diagram of clinical interpretation workflow. Challenges: (b) Current task-specific models lack cognitive faculties, while general MLLMs lack Echo-specific expertise and capabilities. (c) EchoAgent: An agentic system emulating expert interpretation with fully coordinated ``eyes-hands-minds''.} \label{fig1}
\end{figure}

Echocardiography (Echo) stands as a widely used and indispensable non-invasive imaging modality for the assessment of cardiac function, forming the cornerstone of diagnosis, management, and prognostic evaluation across a broad spectrum of cardiovascular diseases~\cite{update2017heart,heidenreich20222022,writing20212020}. The clinical value of Echo, however, is not fully uncovered through the simple observation of raw visual information~\cite{taub2025guidelines,holste2025complete}. Instead, it is unlocked through expert interpretation as shown in Figure~\ref{fig1} (a), which is a complex, integrative task that requires clinicians to orchestrate multi‑dimensional abilities, fundamentally encompassing (1) ``eyes'' for visual observation and view recognition across multiple cardiac views, (2) ``hands'' for precise structure localization and segmentation, as well as quantitative measurement of key parameters, and (3) ``minds'' for learning clinical knowledge, integrating multimodal evidence, and executing logical and reliable diagnostic reasoning. 
\setlength{\textfloatsep}{4pt plus 2pt minus 2pt} 

The pursuit of automating Echo analysis has been a prevalent focus of medical Artificial Intelligence (AI)  research ~\cite{he2023h2former,messaoudi2023cross,deng2024memsam,hu2025echoone,li2023llava,wang2024qwen2,bai2025qwen3,singh2025openai}, primarily advancing along two types of paradigms, as displayed in Figure~\ref{fig1} (b). The first paradigm leverages task-specific deep learning (DL) models~\cite{he2023h2former,messaoudi2023cross,deng2024memsam,hu2025echoone}, which excel in replicating the ``eyes'' and ``hands'' of clinicians, with excellent performance in isolated tasks such as view classification, anatomical structure segmentation, or parameter measurement. Nevertheless, these approaches fall short of performing full-study Echo interpretation as they lack the Echo ``minds'', i.e., the ability to autonomously orchestrate distinct ``hands'' in a clinically-logical order and reason about the clinical significance of their operation results in Echo. Another type of paradigm employs Multimodal Large Language Models (MLLMs)~\cite{li2023llava,wang2024qwen2,bai2025qwen3,singh2025openai}, which exhibit impressive abilities in visual question answering and semantic reasoning, offering a form of ``eyes-minds'' capability. MLLMs can describe image content and answer questions based on visual and textual patterns pretrained from vast datasets. Nonetheless, without specific knowledge ingrained for Echo and the precise ``hands'' for quantitative analysis, these models are hard to fulfill expert-level Echo interpretation as they lack domain-aware expertise and capabilities to obtain crucial quantitative evidence, i.e. Echo-specific ``minds'' and ``hands'', which is often ungrounded and clinically unreliable~\cite{wang2025medagent,fallahpour2025medrax,nath2025vila}. Therefore, we still lack an end-to-end solution with integrated ``eyes‑hands‑minds'' for reliable Echo interpretation.
To tackle these issues, we propose EchoAgent, an agentic system specialized for end‑to‑end Echo interpretation. As illustrated in Figure~\ref{fig1} (c), we aim to design a workflow that emulates the perceptual‑cognitive‑motor process of a cardiac sonographer to learn, observe, operate, and reason. Specifically, EchoAgent accomplishes this through three stages. At the initial stage, our EchoAgent encompasses an essential domain‑aware ``mind'' via a preliminary expertise‑driven cognitive engine. Subsequently, it is operationalized through a hierarchical collaboration toolkit, empowered with ``eyes'' and ``hands''. Ultimately, EchoAgent is capable of executing a fully coordinated ``eyes‑hands‑minds'' workflow under the orchestration of an orchestrated reasoning hub, thus performing reliable reasoning for cardiac function assessment. In short, our contributions can be summarized as follows:
\begin{itemize}
   \item [1)]
    We present a specialized agentic system for Echo interpretation, named EchoAgent, which not only performs reliable Echo analysis within a single framework, but also pioneers a novel ``eyes‑hands‑minds'' paradigm.
   \item [2)]
    To endow EchoAgent with multi‑dimensional capabilities, we design an expertise‑driven cognition engine to build a foundational ``mind'' with domain‑aware knowledge repository construction, followed by a hierarchical collaboration toolkit that functions as ``eyes'' and ``hands'' through diverse perception and operation skills.
   \item [3)]
    Building upon the expertise knowledge base and integrated multimodal information, we further introduce an orchestrated reasoning hub that intelligently collaborates perception, operation, and reasoning  abilities, thereby implementing  stepwise and traceable Echo interpretation.
   \item [4)]
    We conduct comprehensive evaluations of EchoAgent across 48 distinct views covering 14 key cardiac structures. Experimental results demonstrate that EchoAgent achieves state‑of‑the‑art performance, exhibiting an analytical process that closely mimics the ``eyes-hands‑minds'' coordination of human specialists.
\end{itemize}
\section{Related work}
\subsection{Specialized task-specific networks for Echo}
Developing DL networks to automate Echo analysis has become a trendy research objective~\cite{deng2024memsam, hu2025echoone,liu2025q, yang2024cardiacnet,10218329, chao2025foundation}. Plenty of work has focused on specific sub-tasks, primarily including view identification and anatomical structure segmentation. Early approaches employed CNNs to identify views~\cite{khamis2017automatic,zhang2018fully}, while recent advancements have utilized multi-instance learning\cite{song2025echoviewclip} or pretrained foundation models (FMs)~\cite{christensen2024vision, vukadinovic2025comprehensive} to capture complex visual patterns for accurate view prediction. For segmentation, U-Net remains the strong baseline~\cite{leclerc2019deep}, with attention mechanisms being adapted to improve boundary delineation for cardiac structures~\cite{he2023h2former, messaoudi2023cross,painchaud2022echocardiography,zhang2025bridging}. Moving forward, the emergence of FMs have also spurred remarkable advances in Echo segmentation, where FM-based models such as MemSAM~\cite{deng2024memsam} and EchoONE~\cite{hu2025echoone} facilitate precise segmentation for Echo. While these models excel in their designated tasks, they operate with limited ``eyes'' and ``hands'', requiring separate manual assistance for final diagnostic conclusion, which is not favorable for comprehensive Echo interpretation.
\subsection{Advanced MLLMs in medicine}
The advent of MLLMs has introduced a new paradigm for medical image understanding recently. Pre-trained on massive web-scale image-text pairs, pioneers like Qwen-VL~\cite{bai2023qwen,wang2024qwen2,bai2025qwen2}, DeepSeek~\cite{guo2025deepseek,liu2024deepseek,wu2024deepseek}, and OpenAI-GPT~\cite{achiam2023gpt,hurst2024gpt} series exhibit remarkable zero-shot or few-shot capabilities for image captioning and general visual question answering (VQA) in real-world scenarios. Various works have attempted to further pre-train or fine-tune models on biomedical corpora, yielding medical-specialized MLLMs such as LLaVA-Med~\cite{li2023llava} and DeepSeek-R1-Distill-Llama-8B~\cite{hurst2024gpt}. While they demonstrate impressive abilities to interpret and reason semantically, showcasing a basic form of ``eyes-minds'', they remain fundamentally passive and discussion-oriented, given the absence of domain‑specialized expertise and tool‑oriented capabilities to acquire clinical evidence. Consequently, endowing MLLMs with expert ``minds'' and the dexterous ``hands'' for reliable Echo interpretation remains an urgent and necessary challenge.
\subsection{Towards domain-expertise agentic system}
Recent days have witnessed growing interest in developing autonomous agentic systems, driven by the unparalleled tool-calling capability of MLLMs most recently~\cite{bai2025qwen3, singh2025openai}. This promotes the development of agents to transcend the conversational nature of standard MLLMs by actively orchestrating external tools to acquire quantitative evidence and execute complex workflows~\cite{fallahpour2025medrax, nath2025vila,chen2025multi,daghyani2025echoagentguidelinecentricreasoningagent}. For instance, ~\cite{fallahpour2025medrax} introduces a pipeline MedRAX, integrating specialized CXR tools with MLLMs to address complex chest X-ray analysis. In the more generalized CT and X‑ray domains,~\cite{nath2025vila} enhances diagnostic performance by explicitly infusing MLLMs with knowledge from domain‑specialized vision models. Meanwhile,~\cite{chen2025multi} presents a collaborative‑learning framework that adaptively incorporates pre-trained single-modal medical large models with small multimodal models to boost multimodal diagnosis. Despite these advances, we still lack a specialized agent for reliable Echo interpretation, since Echo is a unique imaging modality, characterized by its inherent multi‑view interdependency, complex anatomical geometry, and ambiguous structure boundaries~\cite{daghyani2025echoagentguidelinecentricreasoningagent}. These challenges necessitate a specialized synergy of Echo expertise with perceptual, operational, and reasoning capabilities, which most existing agents are not designed to fulfill.

\section{Method}
\subsection{Overview}
\begin{figure*}[htb]
\vspace{-3ex}
\centering
\includegraphics[width=0.9\textwidth]{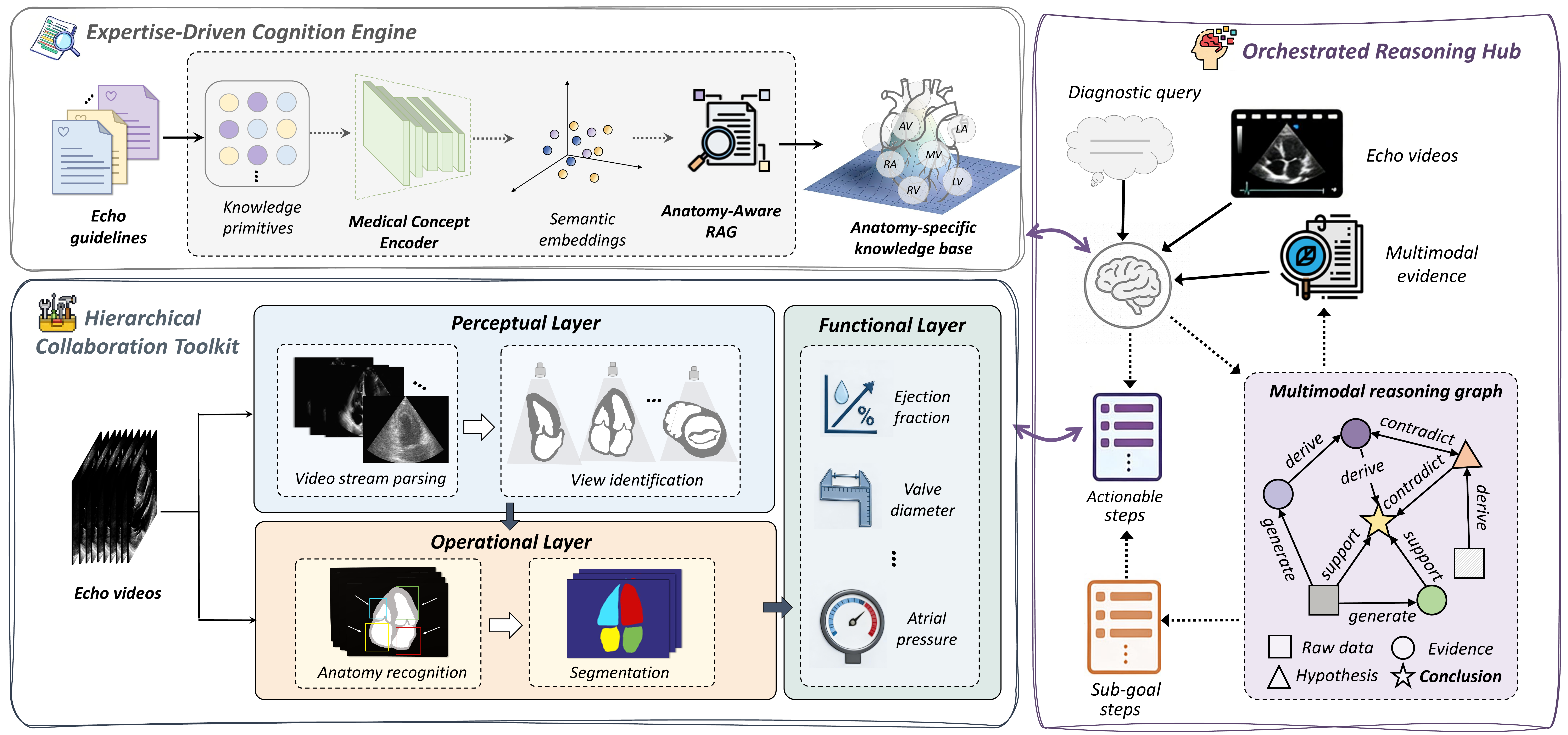}
\caption{Framework of our EchoAgent, which establishes a fully coordinated workflow for Echo interpretation through an expertise‑driven cognitive engine to build an essential Echo ``mind'', a hierarchical collaboration toolkit that delivers manipulation skills, and an orchestrated reasoning hub to dynamically integrate these components, enabling evidence‑based reasoning for cardiac function assessment.} \label{fig2}
\vspace{-2ex}
\end{figure*}

Figure~\ref{fig2} depicts the overall framework of EchoAgent, which comprises three core stages. Mimicking consciousness cultivation of a cardiac sonographer,  EchoAgent firstly formulates an essential Echo-specialized ``mind'' by adapting a general-purpose MLLM (i.e. Qwen3-VL-Plus~\cite{bai2025qwen3}), which assimilates domain expertise in the form of clinical guidelines and medical literature through an expertise‑driven cognition engine, converting heterogeneous textual information into a structured knowledge base. Afterwards, EchoAgent is equipped with requisite operating skills with a powerful combination of FMs within a hierarchical collaboration toolkit, implementing Echo video stream parsing, view identification, anatomical structure recognition and segmentation, as well as diverse quantitative measurements. With ``eyes'', ``hands'' and ``minds'', EchoAgent finally can autonomously perform end-to-end Echo interpretation, with an orchestrated reasoning hub to receives the external clinical diagnostic query and raw Echo videos, which dynamically comprehends multimodal information, orchestrates the toolkit to gather patient-specific evidence, performs structured reasoning grounded to the knowledge repository, and finally draw an evidence‑backed diagnostic conclusion.
\setlength{\textfloatsep}{2pt plus 2pt minus 2pt} 
\subsection{Expertise-driven cognition engine}
As one person needs to learn expertise knowledge to be a qualified echocardiographer, we devise an Expertise‑Driven Cognition (EDC) engine to initialize a specialized ``mind'' of our agent, building a structured knowledge base for 48 cardiac views covering 14 structures of the heart. 

To acquire textual knowledge for Echo, we mainly exploit four expertised sources, which consist of a medical library Unified Medical Language System (UMLS)~\cite{bodenreider2004unified} for general conception learning and specific Echo guidelines from the American Heart Association (AHA), American Society of Echocardiography (ASE) and European Association of Cardiovascular Imaging (EACVI)~\cite{heidenreich20222022,writing20212020,taub2025guidelines,nagueh2025recommendations, grapsa2025diastolic}, which provide credible knowledge for view acquisition, anatomy regions, measurement techniques and diagnostic criteria. Furthermore, we transform heterogeneous domain knowledge into a machine‑actionable repository through the dedicated EDC  engine. First, documents from all sources are preprocessed and decomposed into semantically coherent knowledge primitive $\boldsymbol{P} = \{p_1, p_2, \cdots p_{D}\}$. Each primitive is encoded by a medical concept encoder $f_{\theta}(\cdot)$ and then mapped into a shared high-dimensional embedding space, thereby constructing a hierarchical topology over the knowledge corpus. This embedding process can be formulated as:
\begin{align}
\boldsymbol{e}_i &= f_\theta(p_i), \tag{1}
\end{align}
where $p_i$ denotes the $i$-th knowledge primitive and $\boldsymbol{e}_i$ represents its corresponding semantic representation.

To enable efficient and anatomy-aware knowledge access, we predefine 14 common cardiac anatomical groups $A = \{a_1, a_2, \cdots a_{14}\}$,, including left ventricle, mitral valve, aortic valve and so forth. Each primitive $p_i$ is assigned to one or more anatomical subsets, thus organized into anatomy-indexed subsets:
\begin{align}
\boldsymbol{I}_{a_j} &= \{ p_i \mid a_j \in A(p_i) \}, \tag{2}
\label{eq:two}
\end{align}
Considering a target anatomy $a_t$, the agent first narrows the search to the corresponding subset $\boldsymbol{I}_{a_t}$ by performing anatomy-specific retrieval based on retrieval augmented generation (RAG)~\cite{gao2023retrieval} to extract the $k$ most relevant primitives. The relevance between the target keywords $d_{a_t}$ and a primitive $p_i$ measured by cosine similarity :
\begin{align}
\boldsymbol{sim}(w_{a_t}, p_i) &= \frac{f_\theta(d_{a_t}) \cdot \mathbf{e}_i}{\|f_\theta(d_{a_t})\| \|\mathbf{e}_i\|}. \tag{3}
\label{eq:three}
\end{align}
The top-k primitives with the highest similarity are retrieved:
\begin{align}
P_{top-k} &= \operatorname{top}_k\left( \{ \operatorname{sim}(d_{a_t}, p_i) \mid p_i \in P_{a_t} \} \right). \tag{4}
\end{align}
Based on the retrieved primitives, we conduct structured summary and finally establish a machine-actionable knowledge repository $\boldsymbol{R} = \{\boldsymbol{r}_1, \boldsymbol{r}_2, \cdots, \boldsymbol{r}_{a_t}, \cdots \}$ for all essential cardiac anatomical structures. 

\subsection{Hierarchical collaboration toolkit}
Upon the cultivation of a foundational Echo ``mind'', we further introduce a Hierarchical Collaboration (HC) toolkit to endow EchoAgent with Echo-specific ``eyes'' and ``hands''. The toolkit is assembled by three distinct layers, mirroring the progression from visual observation to manual operation and quantitative functional measurements.

\noindent\textbf{Perceptual layer:} We first design a perceptual layer to process the fundamental visual characteristics of Echo, which is inherently characterized by multi-view interdependency and complex spatiotemporal dynamics. To this end, we harness a dedicated FM, EchoPrime~\cite{vukadinovic2025comprehensive}, to parse the input video streams and automatically identify the diverse echocardiographic views (e.g., apical-2-chamber, apical 4-chamber, parasternal long-axis) present within the sequence. The perceptual layer establishes the essential spatial and anatomical context for subsequent analysis.

\noindent\textbf{Operational layer:} Since the accurate segmentation of specific anatomical structures is a prerequisite for deriving clinically meaningful quantitative metrics, we employ a customized segmentation model that has previously been devised based on USFM~\cite{jiao2024usfm}. This layer performs the automatic delineation of key cardiac structures (e.g., left ventricle,  aorta, right ventricle, left atrium), providing a robust foundation for the following functional analysis.

\noindent\textbf{Functional layer:} Building upon the outputs of the perceptual and operational layers, we further incorporate dedicated fine-tuned versions of USFM and EchoPrime to function as a specialized quantitative analysis engine, which is responsible for calculating critical clinical parameters, such as ejection fraction, chamber dimensions, and right atrial pressure, as well as generating clinical reports to assess function of distinct cardiac structures.

\subsection{Orchestrated reasoning hub}
Having possessed the specialized ``eyes'', ``hands'', and ``mind'', we propose an Orchestrated Reasoning (OR) Hub to specially emulate the neurological system of a cardiac sonographer for reliable and end‑to‑end Echo interpretation, accomplishing a fully coordinated perception-action-reasoning workflow.  
Given a cardiac diagnostic query $Q$ and the corresponding raw Echo videos $V$, the OR Hub initiates a closed‑loop reasoning process tailored for Echo:

\noindent\textbf{1) Expertised Echo knowledge retrieval \& task allocation:} The hub first interacts with the exclusive Echo knowledge repository $\boldsymbol{R}$ stored in the foundational ``mind''. According to Equation~\ref{eq:three}, it performs a latent semantic retrieval to find the the targeted repository $\boldsymbol{R}_{a_q}$ relevant to $Q$: 
\begin{align}
\boldsymbol{R}_{a_q} = \arg\max_{r \in R} \boldsymbol{sim}(f_\theta(Q), \boldsymbol{e}_i). \tag{5}
\end{align}
Subsequently, the hub decomposes $\boldsymbol{r}_{a_q}$ into an actionable sequence of steps $\boldsymbol{S} = \{\boldsymbol{s}_1, \boldsymbol{s}_2, \cdots, \boldsymbol{s}_n\}$, where each step is adaptively scheduled and mapped to the optimal functional tool in the HC toolkit $\boldsymbol{\tau}_{a_q}^* \in T$. Thus, each step $\boldsymbol{s}_i$ can be defined as:
\begin{align}
\boldsymbol{s}_i: \operatorname{Execute}(\boldsymbol{\tau}_{a_q}^*, \operatorname{Input}_i) \tag{6}
\end{align}

\noindent\textbf{2) Dynamic Echo reasoning graph construction:} As each step is executed, the HC Toolkit returns quantitative evidence based on Echo with confidence scores. Rather than forming premature conclusions, the OR Hub incrementally constructs a dynamic multimodal reasoning graph for Echo $G = (N, E)$, in which Nodes ($N$) represent Echo‑specific clinical entities: cardiac diagnostic concepts, execution evidence (with associated confidence scores), and raw data anchors. Edges ($E$) represent Echo‑specific clinical relationships, including: 
    \begin{itemize}
        \item \textbf{Generation:} [Raw Data]--(generates)$\to$[Evidence] (e.g., raw A2C video $\to$ LV segmentation mask).
        \item \textbf{Support/contradiction:} [EvidenceA] $\leftarrow$ (supports /contradicts) $\to$ [Hypothesis/EvidenceB] (e.g., ``EF=33.5\%'' $\longleftrightarrow$ ``Considerably reduced EF/Normal'').
        \item \textbf{Derivation:} [Raw Data/EvidenceA]--(derives)$\to$ [Hypothesis/EvidenceC] (e.g., LV segmentation mask $\to$ LV volume at different phrases $\to$ ``EF=33.5\%'').
    \end{itemize}
    
\noindent\textbf{3) Adaptive Echo reasoning workflow:}  To ensure robust Echo reasoning, the hub performs adaptive inferencing process. Specifically, it firstly considers confidence and evaluates the certainty between cardiac diagnostic hypotheses $H = \{h_1, h_2, \dots\}$ and the current Echo evidence graph $G(t)$, where the posterior probability of each hypothesis is:
\begin{align}
P(h_m | G(t)) \propto P(G(t) | h_m) \cdot P(h_m), \tag{7}
\end{align}
in which the likelihood $P(h_m | G(t))$ is defined by evaluating the consistency of Echo‑specific patterns under hypothesis $h_m$  at time $t$. 

If any step yields low confidence or high uncertainty, the hub triggers an adaptive Echo‑focused mechanism to seek alternative diagnostic pathways, which will actively generate a set of sub-goal $\boldsymbol{S}_{sub}$ like requesting a re‑measurement from a different view. The sub-goal will be inserted into the action step pool $\boldsymbol{S}$. This iterative thinking process persists until the evidence graph achieves a sufficient consistency score or a maximum reasoning depth is reached, which ensures a verifiable and self-reflective reasoning trajectory.

Consequently, we have constructed a fully coordinated ``eyes‑hands‑minds'' workflow, realizing an end-to-end agentic system specifically engineered for Echo. 
\begin{table}[htbp]
\centering
\renewcommand{\arraystretch}{0.9} 
\footnotesize\caption{Dataset statistics}
\label{Table1}
\footnotesize
\resizebox{0.95\linewidth}{!}{
\begin{tabular}{lcccccc}
\toprule
\textbf{Dataset} & \multicolumn{6}{c}{\textbf{CAMUS (EF grading)}} \\
\midrule
\multirow{2}{*}{\begin{tabular}[c]{@{}c@{}}Overall\\statistics\end{tabular}} 
& \multicolumn{2}{c}{Subject number} 
& \multicolumn{2}{c}{Video number} 
& \multicolumn{2}{c}{Frame number} \\
\cmidrule{2-7}
& \multicolumn{2}{c}{500} 
& \multicolumn{2}{c}{1000} 
& \multicolumn{2}{c}{9,268} \\
\midrule
\multirow{3}{*}{\begin{tabular}[c]{@{}c@{}}Specific\\cases\end{tabular}} 
&\multicolumn{2}{c}{\begin{tabular}{@{}c@{}}Normal\\(EF$\geq$50\%)\end{tabular}}
&\multicolumn{2}{c}{\begin{tabular}{@{}c@{}}Mildly reduced\\(40\%$\leq$EF$<$50\%)\end{tabular}}
&\multicolumn{2}{c}{\begin{tabular}{@{}c@{}}Considerably reduced\\(EF$<$40\%)\end{tabular}}\\
\cmidrule{2-7}
& \multicolumn{2}{c}{178} 
& \multicolumn{2}{c}{164}
& \multicolumn{2}{c}{158} \\
\midrule
\toprule
\textbf{Dataset} & \multicolumn{6}{c}{\textbf{MIMIC-EchoQA (Multi-option question-answer)}} \\
\midrule
\multirow{2}{*}{\begin{tabular}[c]{@{}c@{}}Overall\\statistics\end{tabular}} 
& \multicolumn{2}{c}{Subject number} 
& \multicolumn{2}{c}{Video number} 
& \multicolumn{2}{c}{Frame number} \\
\cmidrule(lr){2-7}
& \multicolumn{2}{c}{622} 
& \multicolumn{2}{c}{622} 
& \multicolumn{2}{c}{51,194} \\
\midrule
\multirow{3}{*}{\begin{tabular}[c]{@{}c@{}}Specific\\cases\end{tabular}} 
&{\begin{tabular}{@{}c@{}}Pericar-\\dium\end{tabular}}
&{\begin{tabular}{@{}c@{}}Aortic\\valve\end{tabular}}
&{\begin{tabular}{@{}c@{}}Mitral\\valve\end{tabular}}
&{\begin{tabular}{@{}c@{}}Ventric-\\cles\end{tabular}}
&{\begin{tabular}{@{}c@{}}Atria\end{tabular}}
&{\begin{tabular}{@{}c@{}}Vessels\\\&Others\end{tabular}}
\\
\cmidrule{2-7}
& 82 & 132 & 87 & 190 & 73 & 58 \\

\bottomrule
\end{tabular}
}
\end{table}
\begin{table*}[htbp]
\vspace{-2ex}
\centering
\footnotesize\caption{Comparitive results (Acc and G-mean, \%) on the single-structure task (EF grading). ``E'', ``H'', and ``M'' represent ``eyes'', ``hand'' and ``minds'', respectively. \checkmark denotes the presence of the capability, while \ding{55} denotes absence, and $\bigstar$ denotes a specialized and coordinated state of presence. \textbf{Bold} numbers indicate the best performance.}
\label{Table2}
\footnotesize
\renewcommand{\arraystretch}{0.95} 
\begin{tabular}{l c *{3}{c} *{6}{>{\centering\arraybackslash}p{1.1cm}}}
\toprule
\multicolumn{2}{c}{\multirow{2}{*}{\textbf{Method}}} & \multicolumn{3}{c}
{\textbf{Capabilities}} & \multicolumn{2}{c}{\textbf{     Normal    }} & \multicolumn{2}{c}{\textbf{Mildly reduced}} & \multicolumn{2}{c}{\textbf{Considerably reduced}} \\
\cmidrule(lr){3-5} \cmidrule(lr){6-7} \cmidrule(lr){8-9} \cmidrule(lr){10-11}
\multicolumn{2}{c}{} & H & M & E-H-M & Acc & G-mean & Acc & G-mean & Acc & G-mean \\
\toprule
\multirow{4}{*}{\begin{tabular}[c]{@{}l@{}}Task-specific\\models\end{tabular}} 
& OnimiaNet & \checkmark & \ding{55} & \ding{55} & 74.00 & 64.81 & 74.00 & 71.76 & 78.00 & 83.62 \\
& H2former & \checkmark & \ding{55} & \ding{55} & 74.00 & 59.16 & 65.00 & 58.57 & 63.00 & 72.00 \\
& MemSAM & \checkmark & \ding{55} & \ding{55} & 73.00 & 52.29 & 53.00 & 46.06 & 60.00 & 69.44 \\
& EchoONE & \checkmark & \ding{55} & \ding{55} & 74.00 & 63.54 & 64.00 & 62.32 & 80.00 & 85.05 \\
\midrule
\multirow{4}{*}{\begin{tabular}[c]{@{}l@{}}General-purpose \\ models\end{tabular}} 
& LLaVA-Med & \ding{55} & \checkmark & \ding{55} & 48.00 & 47.43 & 58.00 & 43.05 & 58.00 & 39.97 \\
& Qwen2.5-7B-VL & \ding{55} & \checkmark & \ding{55} & 39.00 & 12.58 & 58.00 & 0.00 & 79.00 & 0.00 \\
& Deepseek-VL2 & \ding{55} & \checkmark & \ding{55} & 40.00 & 12.75 & 60.00 & 0.00 & 80.00 & 21.82 \\
& GPT-5 & \ding{55} & \checkmark & \ding{55} & 44.00 & 44.72 & 61.00 & 0.00 & 55.00 & 52.08 \\
\midrule
\multirow{2}{*}{\begin{tabular}[c]{@{}l@{}}``E-H-M'' \\ workflows\end{tabular}} 
& GPT-5* & \checkmark & \checkmark & \checkmark & 78.00 & 69.52 & 69.00 & 69.04 & 89.00 & \textbf{92.78} \\
& \textbf{EchoAgent} & $\bigstar$ & $\bigstar$ & $\bigstar$ & \textbf{88.00} & \textbf{83.87} & \textbf{80.00} & \textbf{80.16} & \textbf{92.00} & 89.60\\
\bottomrule
\end{tabular}
\vspace{-1ex}
\end{table*}
\setlength{\textfloatsep}{2pt plus 2pt minus 2pt} 
\begin{figure}[htb]
\centering
\includegraphics[width=0.95\columnwidth]{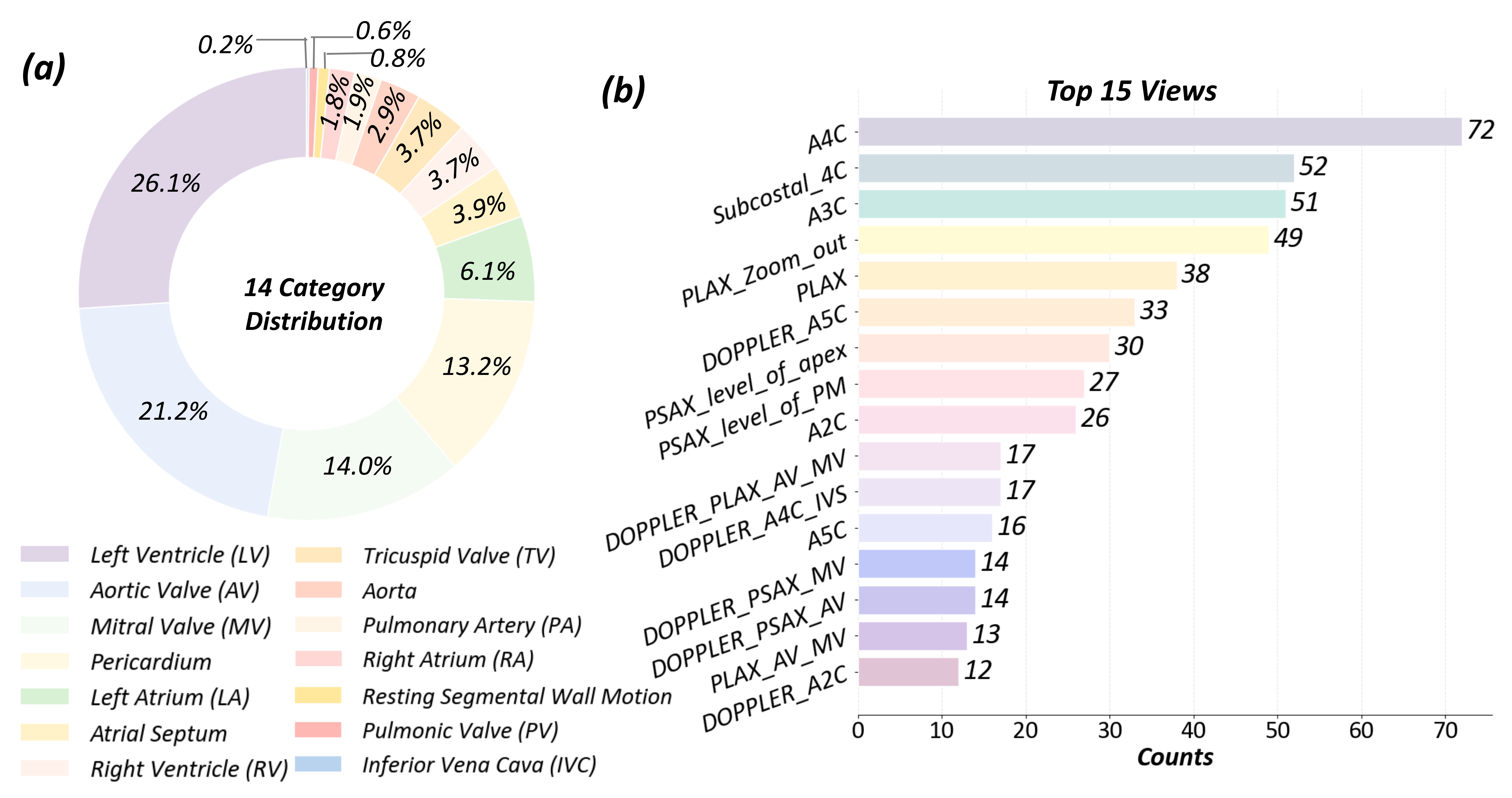}
\footnotesize\caption{Detailed distribution of (a) cardiac structures and (b) top 15 views in the MIMIC-EchoQA dataset.} \label{fig3}
\end{figure}
\section{Experiments and analysis}
\subsection{Experiment setup}
\noindent\textbf{Datasets:} We conduct experiments on two datasets CAMUS~\cite{leclerc2019deep} and MIMIC‑EchoQA~\cite{thapamimic}, which present progressively challenging settings. The CAMUS dataset contains apical-2-chamber (A2C) and apical-4-chamber (A4C) Echo videos with annotations for left ventricle (LV) and labels for ejection fraction (EF), which is essential to assess heart failure~\cite{heidenreich20222022,taub2025guidelines}. The MIMIC‑EchoQA benchmark is built from the MIMIC-IV-ECHO, encompassing 48 distinct views and covering 14 cardiac structures (grouped to 7 major categories), with multiple-choice clinical questions concerning assessment for structures such as pericardium, atria as well as ventricles, vessels and valves. The details of the datasets are summarized in Table~\ref{Table1} and Figure~\ref{fig3}.

\noindent\textbf{Evaluation metrics:} For the EF grading task on the CAMUS dataset, we follow clinical guidelines~\cite{lang2015recommendations,taub2025guidelines} to categorize cases into three grades. The overall classification performance is evaluated by the commonly used accuracy (Acc), and geometric mean (G‑mean) to account for potential class imbalance. Moreover, we perform an area under the receiver operating characteristic curve (AUROC) analysis on EF to assess the quantification capability of those models with ``hands''. For the MIMIC‑EchoQA benchmark, where tasks are formulated as multiple‑choice questions, we present the Acc indicator as detailed per‑group Acc as well as the overall Acc across all questions.

\setlength{\textfloatsep}{3pt plus 2pt minus 3pt} 
\begin{figure}[htbp]
\vspace{-1ex}
\centering
\includegraphics[width=\columnwidth]{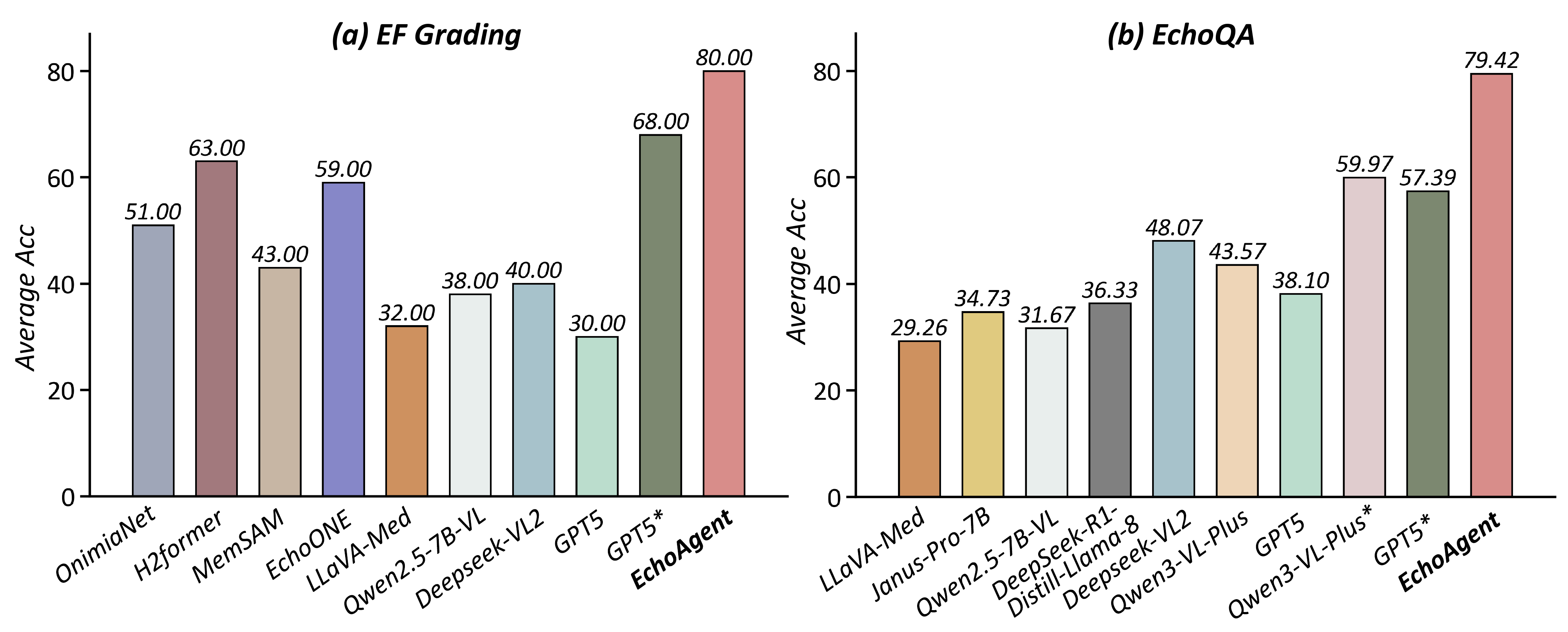}
\caption{Average Acc in (a) EF grading and (b) EchoQA task.} \label{fig4}
\end{figure}
\noindent\textbf{Implementation details:} We implement EchoAgent using PyTorch and LangChain~\cite{topsakal2023creating} to orchestrate the agentic workflow. All experiments are conducted on a server equipped with an NVIDIA GeForce Tesla A100 GPU to ensure efficient finetuning of FMs and inference of MLLMs. For the base ``mind'', we employ Qwen3-VL-Plus~\cite{bai2025qwen3} as the backbone MLLM, which provides fundamental ``eyes-minds'' capabilities for processing general multimodal information and potential learning. In addition, we adhere to the official split ratio of 7:1:2 for training, validation, and testing in the CAMUS dataset, while treating all provided cases as held-out test cases for the MIMIC-EchoQA. Especially, EF in CAMUS is calculated based on segmentation results of LV according to Simpson’s biplane method of disks (SMOD)~\cite{lang2015recommendations}.
\begin{table*}[htbp]
\vspace{-1.5ex}
\centering
\caption{Comparative performance (Acc, \%) of the EchoAgent with state-of-the-art MLLMs on the multi-structure task (EchoQA).}
\footnotesize
\renewcommand{\arraystretch}{0.95}
\label{Table3}
\begin{tabular}{
    c
    *{4}{>{\centering\arraybackslash}p{1.4cm}}
    *{3}{>{\centering\arraybackslash}p{1.3cm}}
}
\toprule
\textbf{Method} &\makecell{Pericardium} & \makecell{Aortic valve} & \makecell{Mitral valve} & \makecell{Ventricles} & \makecell{Atria} & \makecell{Vessels} & \makecell{Others} \\
\midrule
LLaVA-Med & 32.93 & 20.45 & 20.69 & 30.53 & 35.62 & 41.94 & 48.15 \\
Janus-Pro-7B & 39.02 & 26.52 & 27.59 & 40.53 & 30.14 & 51.61 & 37.04 \\
Qwen2.5-7B-VL & 50.00 & 18.18 & 29.89 & 36.84 & 27.40 & 25.81 & 29.63 \\
DeepSeek-R1-Distill-Llama-8B
& 41.46 & 30.30 & 27.59 & 38.42 & 41.10 & 45.16 & 40.74 \\
Deepseek-VL2 & 74.39 & 51.52 & 52.87 & 32.63 & 52.05 & 41.94 & 40.74 \\
Qwen3-VL-Plus & 65.85 & 43.18 & 43.68 & 36.32 & 53.42 & 35.48 & 11.11 \\
GPT-5 & 60.98 & 40.91 & 36.78 & 26.32 & 36.99 & 38.71 & 44.44 \\
\midrule
Qwen3-VL-Plus$^*$ & 73.17 & 56.82 & 56.32 & 57.37 & 69.86 & 48.39 & 51.85 \\
GPT-5$^*$ & 69.51 & 60.61 & 59.77 & 47.89 & 63.01 & 41.94 & 66.67 \\
\textbf{EchoAgent} & \textbf{84.15} & \textbf{82.58} & \textbf{81.61} & \textbf{75.26} & \textbf{80.82} & \textbf{77.42} & \textbf{70.37} \\
\bottomrule
\end{tabular}
\vspace{-2ex}
\end{table*}
%
%
\subsection{Comparison on the single-structure task}
In order to validate the advantages of our method in the sophisticated task of Echo interpretation, we first conduct experiments on the single-structure (i.e., LV) task using the CAMUS dataset to evaluate fundamental cardiac function based on EF grading. Our comparison encompasses three types of paradigms: 
\begin{figure}[htbp]
\vspace{-2ex}
\centering
\includegraphics[width=\columnwidth]{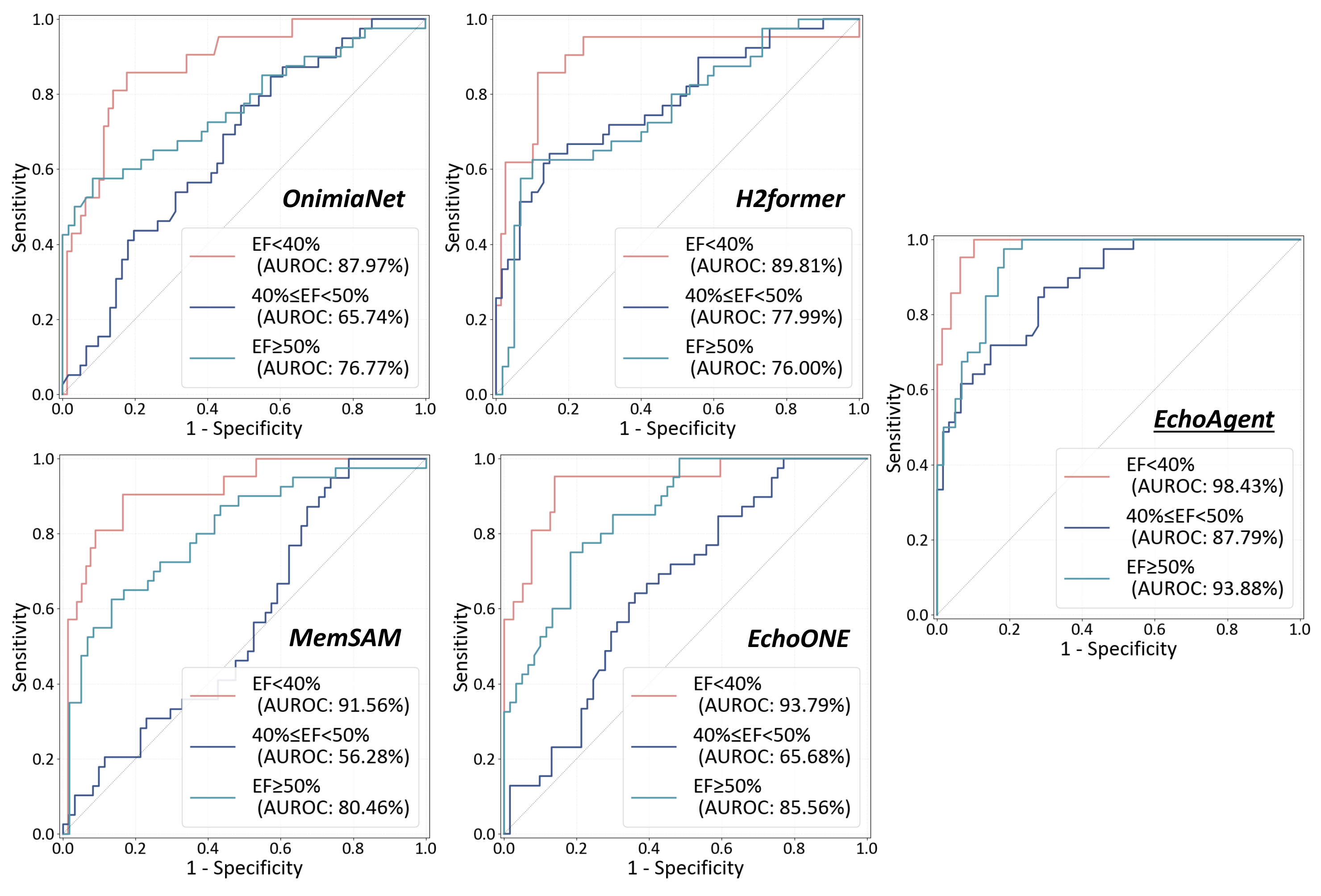}
\caption{ROC curves for LVEF in the CAMUS dataset, derived from segmentation results of models with ``hands''.} \label{fig5}
\vspace{-0.5ex}
\end{figure}
    \begin{itemize}
        \item \textbf{Task‑specific models:} Since LV segmentation is a prerequisite for accurate EF calculation, we evaluate four approaches specifically designed for  Echo segmentation, including two customized networks H2former~\cite{he2023h2former} and OnimiaNet~\cite{messaoudi2023cross}, along with two fine-tuned FMs MemSAM~\cite{deng2024memsam}  and EchoONE~\cite{hu2025echoone}.
        \item \textbf{General‑purpose models:} To assess the capability of current MLLMs for Echo reasoning, we benchmark against four representative models, i.e., LLaVA‑Med~\cite{li2023llava}, Qwen2.5‑7B‑VL~\cite{wang2024qwen2}, DeepSeek‑VL2~\cite{wu2024deepseek}, as well as the latest GPT‑5~\cite{singh2025openai}.
        \item \textbf{``E-H-M'' workflows:} For a more equitable comparison that isolates the benefit of coordination, we adapt the latest baseline GPT-5 which possesses the latent tool-calling capability, into an enhanced GPT-5*, The enhanced configuration equips GPT‑5 with the same ``hands'' (i.e., HC toolkit) used in our proposed pipeline, thus further testing whether simply augmenting a powerful MLLM with tools is sufficient for Echo analysis.
    \end{itemize}
Table~\ref{Table2} and Figure~\ref{fig4} (a) present the quantitative comparison among 10 approaches.
In particular, task‑specific models~\cite{he2023h2former,messaoudi2023cross,deng2024memsam,hu2025echoone} deliver competitive performance with Acc and G-mean exceeding 63.00\% and 58.00\%, respectively, which can be attributed to their specialized capability to operate as skilled ``hands'' for segmentation. However, these models need extra assistance to complete the full diagnostic grading workflow. Instead, general-purpose MLLMs~\cite{li2023llava, wang2024qwen2,wu2024deepseek, singh2025openai} can execute the task autonomously, yet their performance lags behind task‑specific counterparts, with a maximum accuracy drop of 30.00\% and notably low G‑mean scores. This may be due to their lack of fine‑grained perceptual ``hands'', which are critical for complicated Echo interpretation. This highlights the inherent difficulty of precisely interpreting Echo using models with only limited or isolated capabilities. 

Hence, we enhance GPT-5 into GPT-5*, which possesses a basic coordinated workflow. GPT-5* yields a notable improvement over the standard GPT‑5, increasing the average Acc from 30.00\% to 68.00\%. This confirms that augmenting general‑purpose ``mind'' with specialized Echo ``hands'' is not only beneficial but essential for Echo interpretation. However, its performance remained inferior, particularly in the ``Mildly reduced'' grade with both Acc and G‑mean less than 69.10\%,  indicating that the mere tool access is insufficient for generalized Echo diagnosis. In contrast, the proposed EchoAgent achieves superior performance, surpassing all competitors with an average Acc of 80.00\%. Specifically, EchoAgent attains the highest Acc of 88.0\%, 80.0\%, and 92.0\%, as well as G-mean of 83.87\% , 80.16\% , and 89.60\% across three grades. 

To further assess the operational utility, we conducted a ROC analysis on LVEF at clinical thresholds among those models with ``hands'', as shown in Figure~\ref{fig5}. EchoAgent yields remarkable AUROC scores of 98.43\%, 87.79\%, and 93.88\% for three EF grading thresholds, respectively. These consistently discriminative performances indicate the robust ability to reliably identify patients with systolic dysfunction requiring prompt intervention, while minimizing unnecessary follow‑up examinations. Collectively, these results demonstrate that EchoAgent accomplishes superior performance in LV function assessment, verifying the effectiveness of the ``eyes‑hands‑minds'' coordination.
\setlength{\textfloatsep}{2pt plus 2pt minus 2pt} 
\begin{figure*}[htb]
\vspace{-2ex}
\centering
\includegraphics[width=0.85\textwidth]{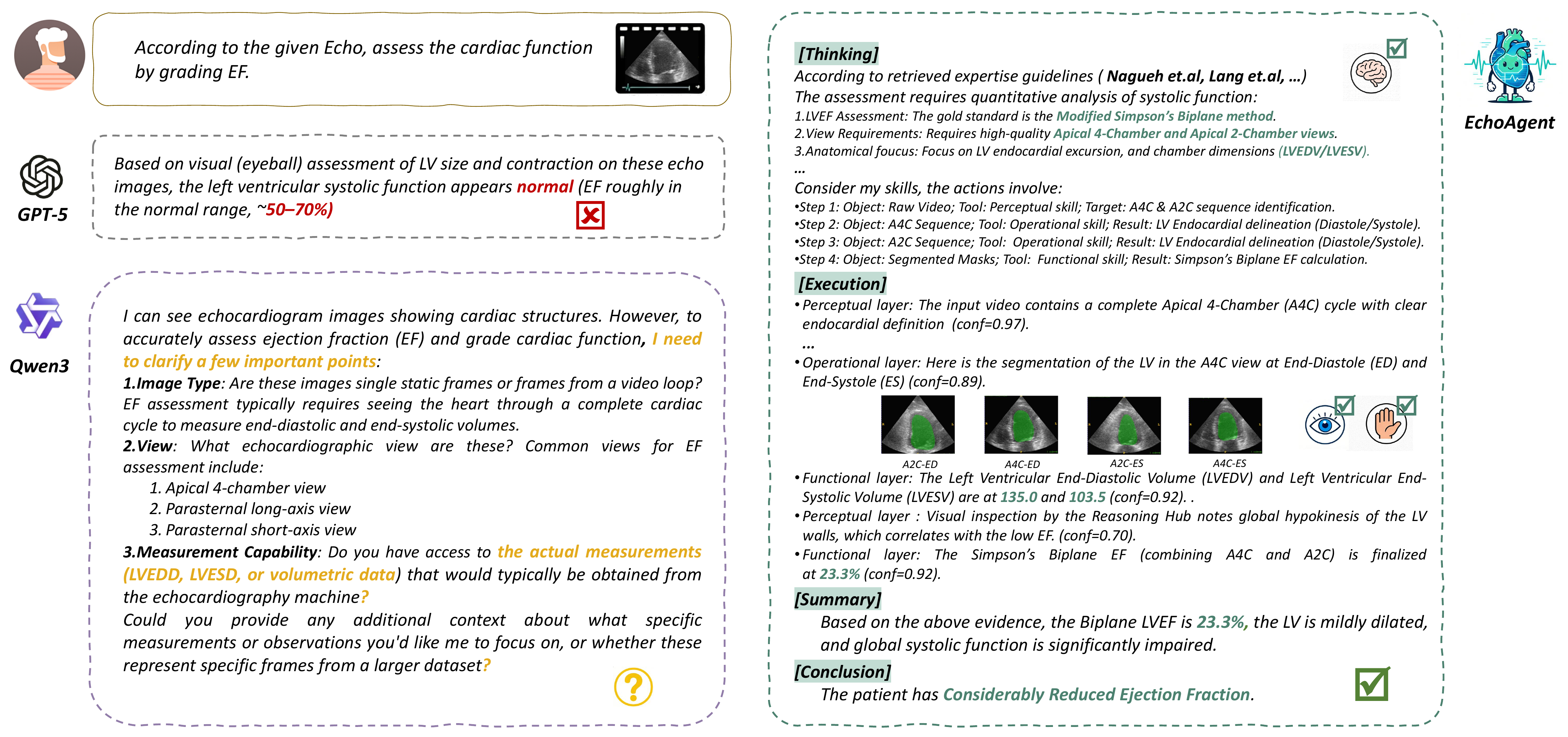}
\caption{A case study for diagnostic analysis of our proposed EchoAgent, compared with two latest MLLMs~\cite{singh2025openai,bai2025qwen3}.} \label{fig6}
\vspace{-2ex}
\end{figure*}
%
%
\subsection{Comparison on the multi-structure task}
To further evaluate the diagnostic breadth and reasoning capability, we carry out experiments on the multi-structure task in the MIMIC-EchoQA dataset, which assesses cardiac function across 7 major anatomical groups with raw Echo videos and clinical question-option pairs. Table~\ref{Table3} and Figure~\ref{fig4} (b) present the comparative results of EchoAgent with various MLLMs.

It is evident that EchoAgent demonstrates balanced and generalized proficiency across all anatomical categories, whereas the representative MLLMs exhibit inconsistent and limited performance. In concrete, pioneers such as LLaVA‑Med~\cite{li2023llava}, Janus‑Pro‑7B~\cite{chen2025janus}, Qwen2.5‑7B‑VL~\cite{wang2024qwen2} and DeepSeek‑R1‑Distill‑Llama‑8B~\cite{guo2025deepseek} provide a baseline level of results and performs relatively well on certain anatomy, obtaining best scores of 48.15\% for  ``Others'', 51.61\% for ``Vessels'', 50.00\% for ``Pericardium'' and 45.16\% for ``Vessels'', respectively. They all struggle on other structures, where corresponding Acc can be as low as 20.45\%, 26.52\%, 18.18\%, and 30.30\% for the clinically-significant cardiac valve ``Aortic valve''. Especially, Deepseek‑VL2~\cite{wu2024deepseek} excels at pericardium,  reaching a score up to 74.39\%. However, its accuracy on the remaining six structural groups remains around 40\%, showcasing an inability to generalize across diverse cardiac anatomy. Moreover, the most recent MLLMs Qwen3-VL‑Plus~\cite{bai2025qwen3} and GPT‑5~\cite{singh2025openai} show moderate improvements in certain categories, yet they still exhibit severe performance limitations. Particularly, GPT‑5 achieves only 26.32\%  on ``Ventricles'', and Qwen3‑VL‑Plus drops to 11.11\% on the ``Others'' category. Nonetheless, when updated to Qwen3-VL‑Plus* and GPT‑5*, their performances show consistent increases of 16.40\% and 19.29\%. This indicates that specialized tools are indispensable for competent Echo interpretation. 

However, even in their enhanced forms, Qwen3-VL‑Plus* and GPT‑5* exhibit notable disadvantages in minor categories, e.g., achieving Acc of only 48.39\% and 41.94\% in ``Vessels''. In comparison, EchoAgent delivers consistently advantageous and balanced performance, with Acc scores above 70.37\% for every anatomical structure. Notably, EchoAgent excels on challenging yet clinically pivotal structures such as the ``Ventricles'', improving Acc by at least 34.73\% over all compared MLLMs. On average, it surpasses the optimal MLLM and latest GPT‑5 by 31.45\%  and 49.36\%, respectively. This superior and stable performance can be attributed to the integrated ``eyes‑hands‑minds'' workflow of EchoAgent, which actively coordinates perception, measurement, and reasoning rather than relying on heuristics or biased attentions. 

In addition, to further validate the merits of the integrated ``eyes-hands-minds'' workflow, we exemplify the inference processes of our EchoAgent with the lately published GPT-5 and Qwen3-VL-Plus in Figure~\ref{fig6}. It can be observed that while general-purpose MLLMs can produce basic thoughts and coherent textual responses for Echo diagnostic questions, their reasoning processes still lack explicit guidance from clinical knowledge and fall short of a traceable chain of evidence. Conversely, EchoAgent provides a structured, evidence-based, and grounded reasoning pathway, which is more favorable in clinical practice.
\vspace{-0.5ex}
\subsection{Ablation studies}
To rigorously evaluate the contribution of three core capabilities, we conduct ablation studies for both tasks. Our experiments begin with a reasoning baseline implemented on Qwen3-VL-Plus, which possesses the fundamental ability to process general multimodal information. We then sequentially investigate the importance of specialized components by: 1) adding the EDC engine to study the effect of a domain‑specialized ``mind'', 2) the HC toolkit to assess the contribution of skilled ``hands'', and 3) integrating both the EDC engine and the HC toolkit under the coordination of the OR hub to construct the complete EchoAgent agent. The overall accuracy is reported in Table~\ref{Table4}.

\noindent\textbf{The necessity of a specialized ``mind'':} The baseline delivers limited Acc scores of less than 45\%. By instilling a specialized ``mind'' through the EDC engine, the model exhibits marked performance improvements, with an increase of 15.00\% for EF grading and 7.88\% for the EchoQA task. These results confirm that a domain‑aware reasoning foundation is essential for moving beyond generic visual understanding toward Echo-specific interpretation.

\noindent\textbf{The power of skilled ``hands'':} Similarly, equipping the system with the HC toolkit also brings enhancement. Compared to the baseline, the HC toolkit raises mean Acc by 37.00\% and 16.40\% for the two tasks separately. In particular, the rise in EF Grading is notable, which may be owing to the fact that EF calculation is inherently dependent on skilled operation ability to perform accurate anatomy segmentation in clinical. These results underscore the significance of fine‑grained operational skills, as reliable cardiac analysis necessitates quantitative evidence.
\begin{table}[htbp]
  \centering
  \caption{Effectiveness (average Acc,\%) of three core components.}
  \label{Table4}
  \footnotesize
  \begin{tabular}{l c c c c c}
    \toprule
    \textbf{Configuration} & \textbf{E} & \textbf{H} & \textbf{M} & \textbf{EF Grading} & \textbf{EchoQA} \\
    \midrule
    Baseline                     & $\checkmark$ & \ding{55}    & $\checkmark$ & 35.00      & 43.57  \\
    Baseline+EDC          & $\checkmark$ & \ding{55}    & $\bigstar$      & 50.00      & 51.45  \\
    Baseline+HC          & $\checkmark$ & $\checkmark$& \ding{55}     & 73.00      & 59.97  \\
    \midrule
    \multirow{2}{*}{\begin{tabular}{@{}c@{}}
    \textbf{Baseline+EDC}\\ 
    \textbf{+HC+OR}
    \end{tabular}} 
    & \multirow{2}{*}{$\bigstar$}  
    & \multirow{2}{*}{$\bigstar$} 
    & \multirow{2}{*}{$\bigstar$} 
    & \multirow{2}{*}{\textbf{80.00}} 
    & \multirow{2}{*}{\textbf{79.42}} \\
    & & & & & \\
    \bottomrule
  \end{tabular}
\end{table}

\noindent\textbf{The synergy of integrated ``eyes-hands-minds'':} When the EDC engine (specialized ``mind'') and the HC toolkit (skilled ``hands'') are incorporated and dynamically coordinated by the OR hub, EchoAgent accomplishes the optimal performance, reaching overall accuracy of up to 80.00\%. It significantly outperforms configurations that possess only the EDC engine or only the HC toolkit, with maximal improvements of 45.00\% and 35.85\% for the EF grading task and knowledge‑intensive EchoQA task, respectively. These demonstrate that the orchestrated synergy of ``eyes-hands-minds'' plays an indispensable role in the reliable end‑to‑end Echo interpretation.
%
%

\section{Conclusion}
In this paper, we propose EchoAgent, an agentic system to emulate the perceptual‑cognitive‑motor process of a cardiac sonographer to learn, observe, operate, and reason, which establishes a reliable and traceable way from raw Echo data to comprehensive Echo interpretation. To achieve this, we introduce an expertise-driven cognitive engine, a hierarchical collaboration toolkit, and an orchestrated reasoning hub to equip the agent with ``eyes-hands-minds''. EchoAgent highlights the need to think how to adapt the ``eyes-hands-minds'' coordination process of humans to an agent for automated and reliable Echo interpretation. We believe EchoAgent can serve as a practical, trustworthy, and powerful assistant to enhance the clinical workflow of Echo diagnosis. 
\section*{Acknowledgements}
This work was supported by National Natural Science Foundation of China (Grant NO. 62531004) and National Key R\&D Program of China (2024YFF0507303).

\small\bibliographystyle{ieeenat_fullname}
\bibliography{main}
\end{document}